\documentclass[letterpaper, 10 pt, conference]{ieeeconf}  % Comment this line out if you need a4paper

\IEEEoverridecommandlockouts                              % This command is only needed if 
\usepackage{amsmath} % assumes amsmath package installed
\usepackage{amssymb}  % assumes amsmath package installed
\usepackage{hyperref}
\usepackage{graphicx}
\usepackage{color,soul}
\usepackage{tabularx}
\usepackage{subcaption}

\title{\LARGE \bf
Guided Deep Reinforcement Learning for Articulated Swimming Robots
}

\author{Jiaheng Hu$^{1}$ and Tony Dear$^{1}$% <-this % stops a space
\thanks{$^{1}$Computer Science Department, Columbia University, New York, NY 10027, USA
        {\tt\small \{jh3916, tbd2115\}@columbia.edu}}%
}

\begin{document}

\maketitle
\thispagestyle{empty}
\pagestyle{empty}

%%%%%%%%%%%%%%%%%%%%%%%%%%%%%%%%%%%%%%%%%%%%%%%%%%%%%%%%%%%%%%%%%%%%%%%%%%%%%%%%
\begin{abstract}

Deep reinforcement learning has recently been applied to a variety of robotics applications, but learning locomotion for robots with unconventional configurations is still limited. Prior work has shown that, despite the simple modeling of articulated swimmer robots, such systems struggle to find effective gaits using reinforcement learning due to the heterogeneity of the search space. In this work, we leverage insight from geometric models of these robots in order to focus on promising regions of the space and guide the learning process. We demonstrate that our augmented learning technique is able to produce gaits for different learning goals for swimmer robots in both low and high Reynolds number fluids.

\end{abstract}

%%%%%%%%%%%%%%%%%%%%%%%%%%%%%%%%%%%%%%%%%%%%%%%%%%%%%%%%%%%%%%%%%%%%%%%%%%%%%%%%
\section{INTRODUCTION}

Articulated swimming robots have attracted much interest from researchers due to their effective locomotive capabilities as well as the richness of their geometric structure. The basis of their locomotion arises from the interaction between actuation of their joints and the surrounding fluid environment. Such interactions depend highly on the nature of the fluid, but previous work has shown that in the cases of extremely low or extremely high Reynolds number fluids, a kinematic system can be approximated, leading to great insights into trajectory planning \cite{hatton2013geometric}.

Even for these idealized systems, however, it is still difficult to derive optimal trajectories analytically. These difficulties are compounded when dealing with robots with more complex morphologies or higher-dimensional joint spaces. Deep reinforcement learning (RL) has recently shown promise to be an effective search strategy, as algorithms have developed to make techniques feasible on physical systems. However, the heterogeneity of the search space and the sparsity of the corresponding reward functions introduce additional challenges for motion planning with RL.

In this paper, we exploit the geometric structure of three-link swimmer systems in low and high Reynolds number fluids to restrict the search space of our reinforcement learning algorithm and learn effective locomoting gaits from a blank slate. We show that this approach is able to speed up training time, as the robot is less likely to be trapped into executing suboptimal gaits. At the same time, we show that the RL method is still flexible enough to be optimized for different objectives, such as energy and speed.

To the best of our knowledge, this is the first attempt to confine RL policy search by utilizing the geometry of the system at hand. This is also one of the first attempts to the locomotion problem of articulated swimmers using model-free deep reinforcement learning.

\section{PRIOR WORK}
\subsection{Geometric Structure}
\begin{figure}[t]
\centering
\includegraphics[width=0.9\columnwidth]{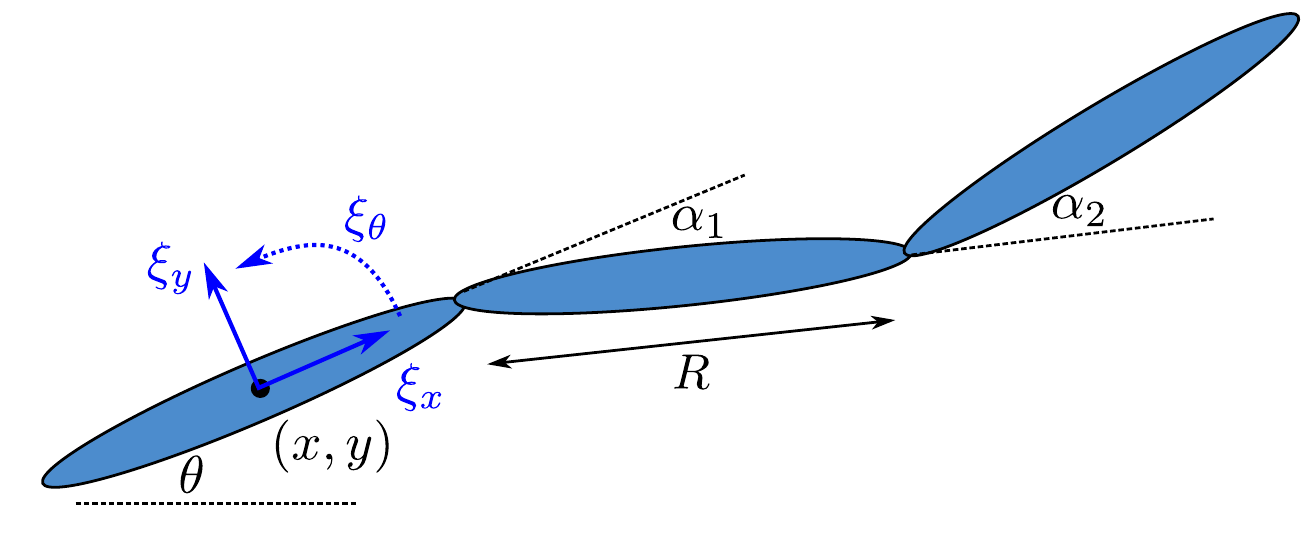} 
\caption{A swimming snake robot comprised of three articulated slender bodies. The coordinates $(x,y,\theta)$ denote the $SE(2)$ inertial configuration of the proximal link, which also has velocities $(\xi_x,\xi_y,\xi_\theta)$ relative to a body-fixed frame. The relative angles of the joints are denoted $\alpha = (\alpha_1,\alpha_2)$.}
\label{fig:swimming}
\end{figure}

In recent decades, techniques and methods from geometric mechanics have been a popular way to model and control mechanical systems. A key idea is that of \emph{symmetries} in a system's configuration space, which allow for the \emph{reduction} of the equations of motion to a simpler form. This has been addressed for general mechanical systems by \cite{marsden2013introduction}, as well as nonholonomic systems by \cite{bloch1996nonholonomic}. For locomoting systems, geometric reduction is often leveraged in tandem with a decomposition of the configuration variables into actuated shape variables and unactuated position variables. If such a splitting is possible, then the configuration space often takes on a \emph{fiber bundle} structure, whereby a mapping called the \emph{connection} relates trajectories between each subspace. Analysis of the connection can then give us intuition into ways to perform motion planning and control of the system, as detailed by \cite{kelly1996geometry} and \cite{ostrowski1998geometric}. This mathematical structure also lends itself to visualization and design tools, detailed by \cite{Hatton01072011}.

Much of the progress in the geometric mechanics of locomotion is predicated on the assumption that the symmetries of a system coincide exactly with the position degrees of freedom. Robots that can be modeled with nonholonomic constraints are examples in which these symmetries occur. Nonholonomic wheeled snake robots have received considerable attention from researchers such as \cite{ostrowski1999computing} and \cite{Shammas01102007a} treating them as \emph{kinematic} systems, so named because constraints that eliminate the need to consider second-order dynamics when modeling its locomotion. This allows for the treatment of the system's locomotion, and subsequent motion planning, as a result of geometric phase (see \cite{murray1993nonholonomic, mukherjee1993nonholonomic, kelly1995geometric, bullo2001kinematic}).

Geometric methods have also examined systems locomoting in fluids. As with terrestrial systems, such a description is most useful if the position degrees of freedom correspond to system symmetries and the rest to internal shape. For single bodies, motion may be achieved through temporal deformation of the body's shape. For articulated swimmers like the three-link robot shown in Fig.~\ref{fig:swimming}, deformation occurs naturally when joints are moved relative to each other (see \cite{hatton2013geometric, melli2006motion, burton2010two}), analogous to the terrestrial version of the system.

Articulated swimmer robots belong to a family of general snake-like robots, which are characterized by a large number of degrees of freedom and locomotion patterns that exhibit cyclic motions through coordination of their joints. Therefore, snake-like robots are usually controlled through kinematics-based methods \cite{ma_icra, tesch2009parameterized}. These methods, however, often rely on hand-tuning a number of different parameters, which can be costly as well as inflexible in new environments.

\subsection{Gait Optimization and Reinforcement Learning}

The problem of gait optimization has been approached through a variety of traditional optimization methods, such as evolutionary algorithms \cite{evolutionary_chernova}, gradient-based methods \cite{Kohl2004MachineLF} and Bayesian optimization \cite{bayesian_calandra}. However, these methods often suffer from local optima, and while the resulting gaits appear effective in locomoting the robots, they are often still quite inefficient when compared to the natural motion achieved by animals. 

Reinforcement learning is a data-driven method that searches for a reward-maximizing policy under a given environment. As an algorithm based on trial-and-error, it has the advantage of not requiring a specific model of the environment or expert knowledge of the problem. With recent advancements in deep neural network and reinforcement learning algorithms, reinforcement learning has become a useful tool for solving robot control tasks such as walker's locomotion \cite{2017-TOG-deepLoco}, dexterous manipulation \cite{DBLP:journals/corr/abs-1709-10087}, and autonomous driving \cite{long2017optimally}. 

There have been a few attempts to solve the problem of gait optimization through reinforcement learning. Bing et al.~\cite{bing2019energyefficient} used PPO to train a forward-locomotion controller for a wheeled snake robot and were able to generate gaits that out-perform those derived from Bayesian optimization and grid search. Sharma and Kitani \cite{Sharma2018PhaseParametricPF} proposed phase-DDPG, where they explicitly trained a cyclic policies for a walker robot by oscillating the weight of the policy network with the phase of the robot. These methods were able to generate fairly natural gaits on certain robots, but often failed to converge to global optima as the robot environment grew more complex. For example, none of the methods were able to solve the swimmer environment \cite{swimmer_remi}.

\section{MODEL AND METHODS}
\subsection{Swimmer Model}
As shown in Fig.~\ref{fig:swimming}, our swimmer robot consists of three rigid links, each of length $R$, which can rotate relative to one another. Its configuration is defined by $q \in Q = G\times B$, where $g = (x,y,\theta)^T \in G = SE(2)$ specifies the position and orientation of the first link in an inertial frame; we measure a link's position at the center of the link. The joint angles $\alpha = (\alpha_1, \alpha_2)^T \in B$ specify the links' relative orientation. We can view $Q$ as a principal fiber bundle, in which trajectories in the shape or base space $B$ lift to trajectories in the group $G$ (see \cite{kelly1995geometric}).

\subsubsection{Low Reynolds Number} Following the treatment of \cite{hatton2013geometric}, we assume that the swimmer is comprised of three slender bodies and suspended in a planar fluid. In the low Reynolds number case, viscous drag forces dominate inertial forces. This allows us to approximate the drag forces as linear functions of the system's body and shape velocities $\xi$ and $\dot \alpha$; we also assume that net forces acting on the system are zero for all time due to damping out by drag forces. We can then derive a \emph{Pfaffian constraint} on the swimming system's velocities as 
\begin{equation}
    F = \begin{pmatrix} F_x \\ F_y \\ F_\theta \end{pmatrix} = \begin{pmatrix} 0 \\ 0 \\ 0 \end{pmatrix} = \omega_1(\alpha) \xi + \omega_2(\alpha) \dot \alpha,
\label{eq:reconstruction}
\end{equation}
where $\omega_1 \in \mathbb R^{3\times3}$ and $\omega_2 \in \mathbb R^{3\times2}$.  The variables $\xi = (\xi_x, \xi_y, \xi_\theta)^T$ give us the body velocity of the system, as shown in Fig.~\ref{fig:swimming}. In $SE(2)$, the mapping that takes body velocities to inertial velocities is given by $\dot g = T_e L_g \xi$, where
\begin{equation}
T_e L_g = \begin{pmatrix} \cos \theta & -\sin \theta & 0 \\ \sin \theta & \cos \theta & 0 \\ 0 & 0 & 1 \end{pmatrix} .
\label{eq:TeLg}
\end{equation}

The full forms of these components can be found in \cite{hatton2013geometric}. The general approach would be to first compute local drag forces on each link, and then combine them to find the total force components for each of the body frame directions. In addition to the system link length $R$, the kinematics also utilize the drag constant of the fluid, characterized by $k$. Since the number of independent constraints is equal to the dimension of the group, these equations are sufficient to derive a kinematic connection for the system (\cite{Shammas01102007a}). In other words, the constraint equations fully describe the first-order dynamics of the group variables in terms of the shape variables only. Thus, Eq.~\eqref{eq:reconstruction} can be rearranged to show this explicitly as the \emph{kinematic reconstruction equation}: 
\begin{equation} \xi = -\mathbf A(\alpha) \dot \alpha = -\omega_1^{-1} \omega_2 \dot \alpha. \label{eq:reconstruction} \end{equation}
$\mathbf A(\alpha)$ is called the \emph{local connection form}, a mapping that depends only on the shape variables, in this case $\alpha_1$ and $\alpha_2$. 

\subsubsection{High Reynolds Number} The high Reynolds number case is opposite from the low Reynolds number environment in that inertial forces dominate viscous forces. Despite the entirely different swimming conditions, the model of the swimmer robot can once again be approximated as kinematic. A Lagrangian for the robot can be expressed in terms of its kinetic energy, as there is no means of storing energy or application of external forces:
\begin{equation} L = \frac12 \begin{pmatrix} \xi & \dot\alpha \end{pmatrix} \mathbb M(\alpha) \begin{pmatrix} \xi \\ \dot\alpha \end{pmatrix}. \label{eq:lagrangian1} \end{equation}
The mass matrix $\mathbb M$ is a function of the system configuration $\alpha$, and it can be decomposed into blocks containing the system's local connection \cite{Shammas01102007a}:
$$ \mathbb M(\alpha) = \begin{pmatrix} \mathbb I(\alpha) & \mathbb I(\alpha) \mathbf A(\alpha) \\ (\mathbb I(\alpha) \mathbf A(\alpha))^T & m(\alpha) \end{pmatrix}. $$

To derive the mass matrix $\mathbb M$, we recognize that the Lagrangian of the three-link system is equal to the sum of the Lagrangians $L_i$ of each of the individual links. Each link has an associated inertia tensor $I_i$ dependent on the shape that we use to model it. In addition, each link has an \emph{added mass} $\mathcal M_i$, which arises due to the inertia of a displaced fluid as a body moves through it; like the inertia tensor, $\mathcal M_i$ is solely a function of the body geometry. \cite{hatton2013geometric} gives an example of the added mass tensor for an elliptical body. The total effective inertia of a single link is then $I_i + \mathcal M_i$, which gives us a Lagrangian of the form
\begin{equation} L = \sum_{i=1}^3 L_i = \sum_{i=1}^3 \frac12 \xi_i^T (I_i + \mathcal M_i) \xi_i \label{eq:lagrangian2} \end{equation}

Once the total Lagrangian is written down, it can be rearranged into the form of Eq.~\eqref{eq:lagrangian1}, from which the local connection $\mathbf A(\alpha)$ can then be extracted.

\subsubsection{Connection Visualization} The structure of the connection form in Eq.\ \eqref{eq:reconstruction} can be visualized in order to understand the response of $\xi$ to input trajectories without regard to time, according to \cite{Hatton01072011}. We can first integrate each row of Eq.\ \eqref{eq:reconstruction} over time to obtain a measure of displacement corresponding to the body frame directions. In the world frame, this measure provides the exact rotational displacement, \textit{i.e.}, $\dot \theta = \xi_\theta$ for the third row, and an approximation of the translational component for the first two rows. If our input trajectories are periodic, we can transform this ``body velocity integral'' into one over the trajectory $\psi: [0,T] \rightarrow B$ in the joint space, since the kinematics are independent of input pacing. Stokes' theorem can then be applied to perform a second transformation into an area integral over $\beta$, the region of the joint space enclosed by $\psi$:
\begin{equation}
-\int_0^T \mathbf A(\alpha(\tau)) \dot \alpha(\tau)\, d\tau = -\int_{\psi} \mathbf A(\alpha)\, d\alpha = -\int_{\beta} \text{d} \mathbf A(\alpha).
\label{eq:stokes}
\end{equation}
The integrand in the rightmost integral is the exterior derivative of $\mathbf A$, computed as the curl of $\mathbf A$ in two dimensions. For example, the connection exterior derivative of Eq.\ \eqref{eq:reconstruction} has three components, one for each row $i$ given by
\begin{equation}
\text{d}\mathbf A_i(\alpha) = \frac{\partial \mathbf A_{i,2}}{\partial \alpha_1} - \frac{\partial \mathbf A_{i,1}}{\partial \alpha_2},
\label{eq:component_of_derivative}
\end{equation}
where $\mathbf A_{i,j}$ is the element corresponding to the $i$th row and $j$th column of $\mathbf A$.

\begin{figure}[t]
\centering
\includegraphics[width=0.88\columnwidth]{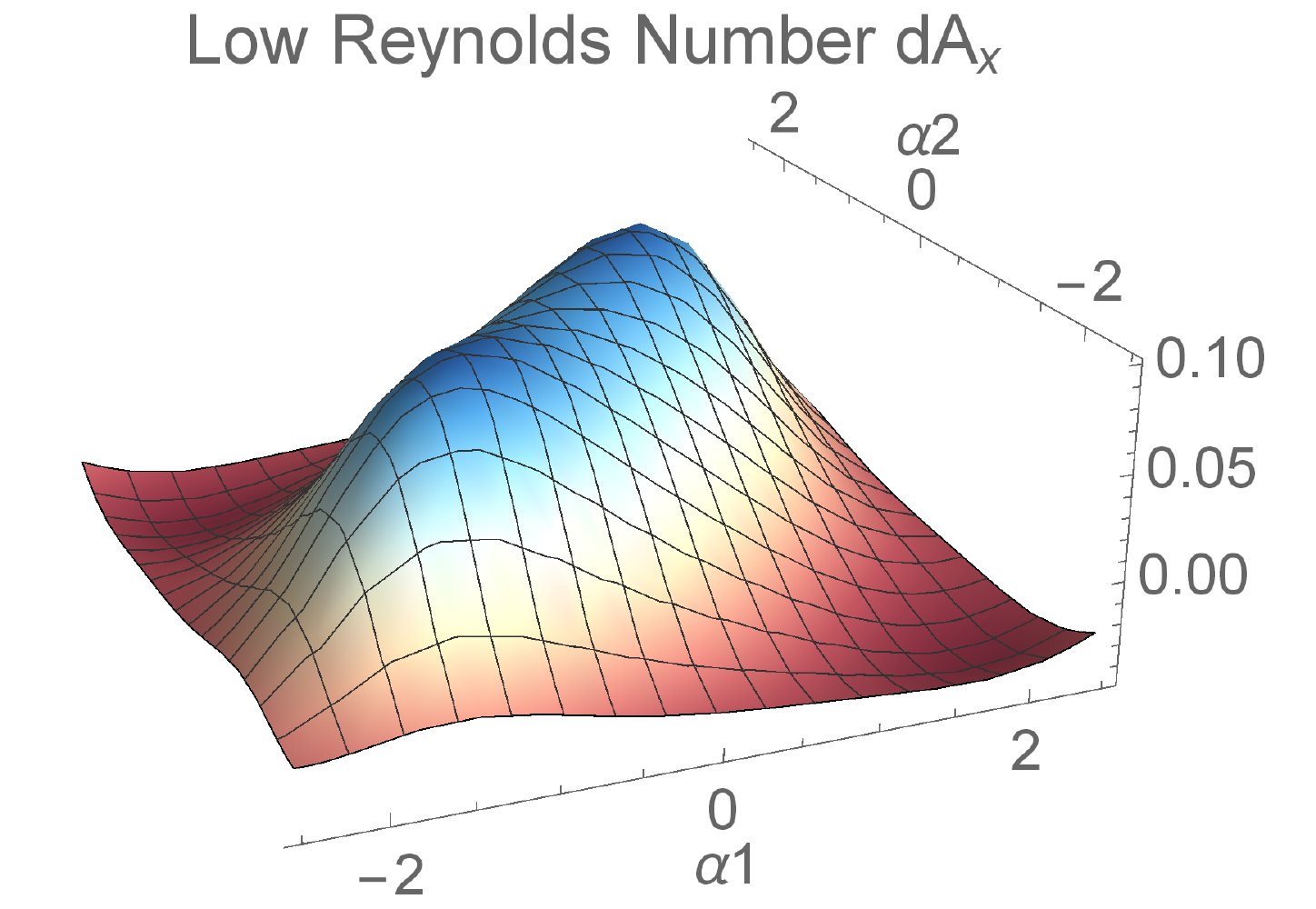}
\includegraphics[width=0.82\columnwidth]{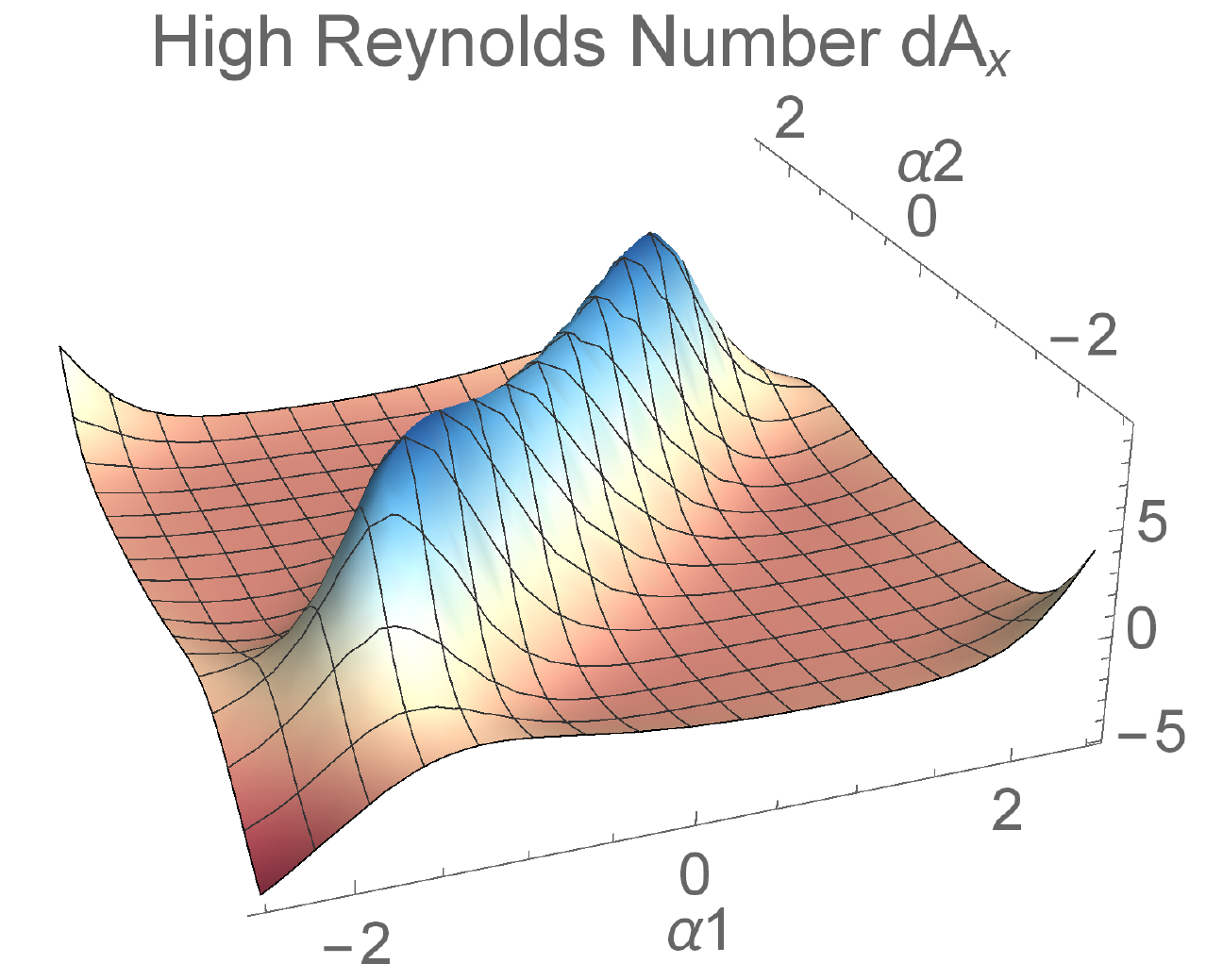}
\caption{Visualizations of the body-$x$ components of the local connection's exterior derivative for the low and high Reynolds swimmers, respectively. Periodic trajectories can be represented as closed curves on these surfaces, and the robot's associated displacement corresponds to the enclosed volume.}
\label{fig:curv-plots}
\end{figure}

The magnitudes of the body-$x$ component (first row) of the connection exterior derivative of each swimmer over the $\alpha_1$-$\alpha_2$ joint space, for a fixed set of sample parameters, are depicted in Fig.~\ref{fig:curv-plots}. The area integral over an enclosed region is the geometric phase, a measure of the expected displacement in the corresponding direction. In particular, a trajectory that advances in a \emph{counter-clockwise} direction over time in joint space will yield positive displacement, since that corresponds to a positive area integral; negative displacement is achieved with a \emph{clockwise} trajectory.

For both swimmers, we see that a high value of the body velocity integral, and thus a high displacement per gait cycle, is generally achieved by executing gaits that encircle a zero contour of these exterior derivative surfaces. However, the optimal  parameters of these gaits differ for the two swimmers, with a larger range for the low Reynolds case and a smaller range for the high Reynolds case. In addition, the means of finding a gait is not obvious when the joint angles are restricted to be smaller than the zero contour. Finally, while we do not show them here we may also be concerned with the $y$ and $\theta$ components as well. Analytically optimizing gaits is thus equivalent to solving a multi-objective constrained optimization problem over a continuous space, a task that becomes exponentially more difficult with increasing system complexity.

\subsection{Baseline-Guided Policy Search (BGPS)}
Based on the geometric models of the robots, we propose an augmented reinforcement learning algorithm called Baseline-Guided Policy Search (BGPS), in which we \textit{restrict} the policy search space of the learning algorithm by utilizing a baseline policy approximated from the geometric structure.

\subsubsection{Robot Environment Setup}
In this work, we focus on locomotion for three-link swimmer robots; the study of more complex robots will the subject of future work. The state of the robot at time $t$ is $s_t = (\alpha_1, \alpha_2, \theta, t)$ , which contains both the joint angles and orientation of the swimmer. The action taken by the robot at time $t$ is $a_t = (\dot\alpha_1, \dot\alpha_2)$, the velocities of the two joints. We investigate two choices of reward functions, which corresponding to two tasks with different optimization goals. 

The first task is to optimize the total distance the robot travels in a pre-determined direction in a given amount of time. The reward is therefore very straightforward: after the robot makes a transition $(s_t, a_t, s_{t+1})$, the value of the reward function $R_t$ is set to be
\begin{equation} R_t = x_{t+1} - x_{t}. \end{equation}  

The second task is to simultaneously maximize the distance travelled and minimize the energy spent. We use a kinetic energy metric and define the reward function as
\begin{equation} R_t = x_{t+1} - x_{t} - \beta \| \dot{\alpha} \|, \end{equation}
where $\beta$ is a coefficient that controls the weight of the energy penalty.

\subsubsection{Proximal Policy Optimization}
A number of reinforcement learning algorithms have been shown to be effective for different physical systems, although the comparison of their various performances is not the focus of this paper. For this work, we choose the proximal policy optimization (PPO) algorithm by Schulman et al.~\cite{schulman2017proximal}, in which an agent seeks to optimize the surrogate objective within the trust region by clipping the probability ratio. PPO has been shown to outperform other online policy gradient methods, with the advantage of being easy to implement. 

\subsubsection{Baseline from Geometric Structure}
The key idea of this work is that we can exploit what we know about the system structure, \textit{e.g.}, as shown in Fig.~\ref{fig:curv-plots}, to help restrict the search space in which reinforcement learning operates. Specifically, the exterior derivative plots suggest that the optimal gaits for moving forward can be roughly approximated as single-frequency sinusoidal functions whose joint-space loops overlay the blue ridges and whose phases are large enough to encircle the widths of the same. Note that the actual optimal policies have no such restriction, \textit{e.g.} as single-frequency sinusoidal functions. This is particularly the case if we have joint limits that prevent the joint angles from extending all the way out to the zero contour at the ends of the ridges. However, such an approximation is sufficient for formulating a \textit{baseline policy} from which RL techniques can then improve upon with a large number of degrees of freedom.

\subsubsection{RL Policy from Baseline}
Once we obtain a baseline policy $ \pi_{base}(s)$ through the method described above, we then use reinforcement learning to search for a separate policy $ \pi_{RL}(s)$. Our eventual policy is then 
\begin{equation} \pi_{final}(s) = \pi_{base}(s) + \pi_{RL}(s) \end{equation}
The most important reason for using a baseline is that we can now control the size of the policy search space by reducing the action range of our RL-learned policy, $|\pi_{RL}|$. By doing so, we limit the policy search to be within the vicinity of our baseline policy, thus guiding the policy search. A properly small action range can shape the policy search space to be near convex, allowing gradient-based methods like RL to be particularly suitable.

\subsubsection{Action Range from Geometric Structure}
Given an environment step length $t$, the amount of deviation $\delta$ that the robot is allowed from the baseline policy, and an action range $\alpha$, we can relate these quantities as $\delta = \alpha t$. Thus, for each action cycle of length $T$, the maximum deviation per cycle is $\delta_{total} = \alpha T = N \delta$, where $N$ is the number of steps per cycle. 

The choice of action range $\alpha$ is another parameter whose value can be informed by the system's geometric structure. $\alpha$ can be interpreted as the maximum amount that we would allow the policy to ``stray'' away from the baseline. Since the baseline is just an approximation for the optimal policy, $\alpha$ needs to be sufficiently large to allow exploration of the policy space to occur. However, the exterior derivative plots can also give us an upper bound on the action range, as there is a finite distance away from our chosen baseline at which the effectiveness of an action would start to drop.

\section{LEARNING AND RESULTS}

\begin{table*}[t]
\begin{subtable}{1\textwidth}
  \centering
  \begin{tabular}{ |c|c|c|c|c|c|c|c|c| }
 \hline
   & BFG & PPO & Phase-DDPG & BGPS (0.6) & BGPS (0.3) & BGPS (0.2) & BGPS (0.15) & BGPS (0.1)  \\
 \hline
 Distance   & 111.05 & 31.79 & 1.14  &  32.08  & 39.39  &  117.6  & \bf 133.3  &  130.8   \\
 \hline
 Energy   & 75.08  & 28.58 &  0.08  &  21.63 & 15.61 & 29.88 & 37.58 & \bf 85.22  \\
 \hline
  \end{tabular}
\end{subtable}

\bigskip
\begin{subtable}{1\textwidth}
  \centering
  \begin{tabular}{ |c|c|c|c|c|c|c|c|c| }
 \hline
   & BFG & PPO & Phase-DDPG & BGPS (0.6) & BGPS (0.3) & BGPS (0.2) & BGPS (0.15) & BGPS (0.1)  \\
 \hline
 Distance  & 94.71 & 19.73  & 13.27  &  122.8  & 116.5  &  \bf 141.8  &  126.2  &  121.4   \\
 \hline
 Energy  & 58.75  & 13.20  &  9.62  &  9.04 & 9.47 & 72.87 & \bf 77.31 & 76.47   \\
 \hline
  \end{tabular}
\end{subtable}
  \caption{The average reward of the learned policy for the low Reynolds swimmer (top) and high Reynolds swimmer (bottom). BFG refers to the baseline policy that we observed from the robots' geometric structures (no learning). PPO and phase-DDPG are the main algorithms to which we compared results. BGPS refers to Baseline-Guided Policy Search (our method), with results provided for several choices of action range for different trials.}
\label{table:results}
\end{table*}

We implement BGPS with different action ranges, and compare the performances directly with PPO and phase-DDPG \cite{Sharma2018PhaseParametricPF}. Our results show that our algorithm generally outperforms the other methods, and that a smaller action range is able to boost the performance of the learned policy, confirming the importance of confining the policy search space.

\begin{figure}[t]
\centering
\includegraphics[width=.85\columnwidth]{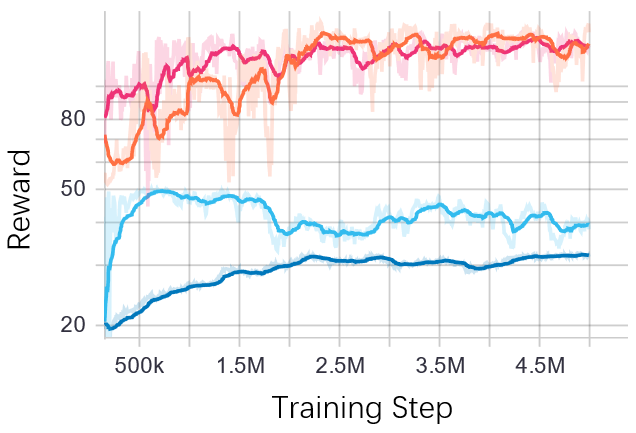} 
\caption{The training curve of different action ranges for optimizing the travelled distance for the low Reynolds swimmer. Red: 0.1, orange: 0.2, cyan: 0.3, blue: 0.6.}
\label{fig:train_curve}
\end{figure}

\subsection{Parameters}
We implemented both a low and high Reynolds three-link swimmer for our simulations. We used a link length of 0.3 m for the low Reynolds case, a nod toward the prevalence of micro-swimmers in this category. For the high Reynolds case, we set the fluid density to $\rho = 1$ kg/m$^3$, and treat the links as ellipses with semi-major axis $a=4$ m and semi-minor axis $b=1$ m. The exterior derivative plots of the swimmers in Fig.~\ref{fig:curv-plots} were obtained using the same parameter values. Our environment step time was set to 0.04 s per step. For both the low and high Reynolds swimmer, we run separate trials for optimizing the speed with and without energy concern. We set $\beta$ to 0.1 for the task of optimizing for energy usage.

\subsection{Network Architecture}
We followed the settings outlined in Schulman et al.~\cite{schulman2017proximal} for implementing PPO. Our policy network, which maps from observation to action, consists of two hidden layers of size 64 and a linear output layer at the end. Rectified Linear units (ReLU) were used as the activation function for every layer except the output layer. Our value network has the same architecture as our policy network, except mapping from (observation, action) to value space. No parameter is shared between the two networks.

\subsection{Training Settings}
We run our experiments on a a computer with an i7-8650U CPU running at 1.90Ghz and an Nvidia GTX 1070 GPU. For each given algorithms and settings, we run for 2.5 million time steps. For each single trial, our algorithm takes about 3 hours to run.

\begin{figure}[t]
\centering
\includegraphics[width=0.98\columnwidth]{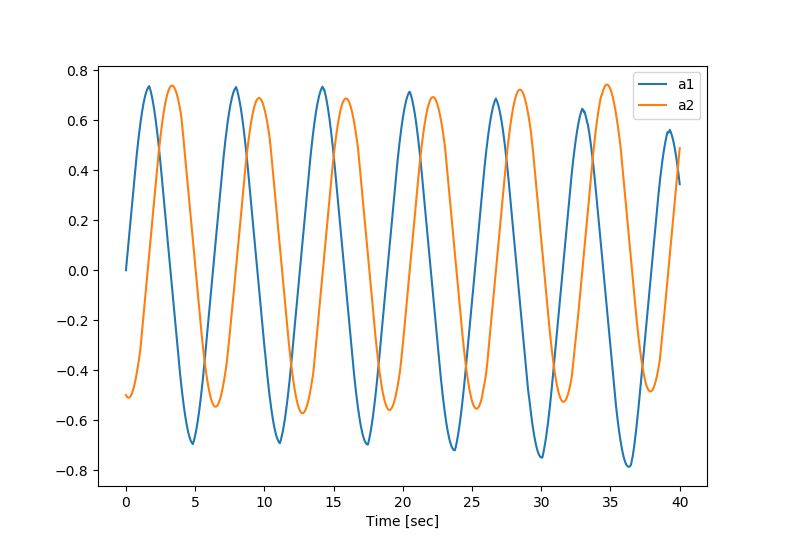} 
\includegraphics[width=0.9\columnwidth]{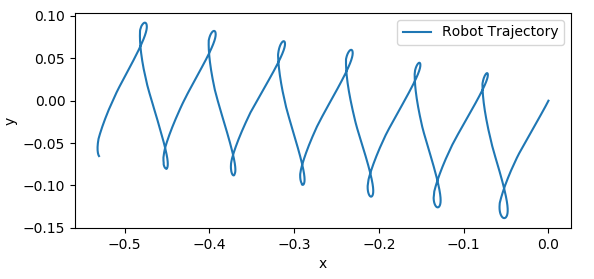} 
\caption{Joint angle (top) and workspace (bottom) trajectories of the low Reynolds swimmer from the best learning trian (BGPS 0.15). The joint angle trajectories are similar to but improve upon the baseline derived from geometry.}
\label{fig:result}
\end{figure}

\subsection{Results}
Table ~\ref{table:results} shows the results of different algorithms for learning locomotive gaits for each swimmer. The "Distance" row refers to the task of maximizing the distance traveled per time in a given direction (the x axis), and the "Energy" row refers to the task of locomoting the robot forward while simultaneously minimizing the energy spent. 

BFG refers to ``baseline from geometry,'' which is the baseline gait we estimated by looking at the geometric model of the robot. For both swimmers, we set a baseline of $0.6cos(t)$ for each joint, with a phase difference of 1 rad between them. Baseline-Guided Policy Search (BGPS) is our method, and the accompanying number on each column header marks the action range for that trial. Both PPO and Phase-DDPG are learning from scratch without utilizing the geometric model, and both of them perform extremely poorly comparing to the other methods shown. In particular, they are unable to learn a gait that performs even close to the baseline gait derived from simple inspection.

BGPS also performs poorly when the action range is too large, but beats all other baselines as the action range is reduced. Fig.~\ref{fig:train_curve} shows the training curve of optimizing the distance for the low Reynolds swimmer. We can clearly see from the plot that training tends to converge to a higher reward when the action range is between 0.1 and 0.2, but fails to converge when between 0.3 and 0.6. This shows that a smaller action range within the appropriate region is the key to our algorithm's success at locomoting the swimmer. For both the low and high Reynolds swimmers, our algorithm produced the best result for both the task of optimizing distance and of minimizing energy spent, among all the methods we tested.

The joint angle and workspace trajectories of the low Reynolds swimmer learned from the best trial (BGPS 0.15) are shown in Fig.~\ref{fig:result}. As expected, the joint angle trajectories are not entirely too different from the baseline that we wrote down from inspection of geometry. However, subtle differences, such as the varying of the relative phases and amplitudes of the two joints over time, suggest the existence of higher-frequency components that were not at all obvious from simple inspection. The accompanying workspace trajectory maximizes the distance reward compared to the other learning trials, as shown in the first row of Table \ref{table:results}.

\section{CONCLUSIONS AND FUTURE WORK}

We have leveraged traditional motion planning techniques from geometric mechanics to make deep reinforcement learning feasible for training articulated swimming robots. Such systems exhibit challenges, such as a policy search space with many local optima, that have previously made it difficult for common DRL approaches. Our approach, which combines intuition with learning, is able to produce superior results for different robot models and different environments.

The fact that our algorithm is able to work across different tasks and robots suggests that this method may easily be generalized. Other robots with similar kinematics or even dynamics can benefit from initialization with an informed baseline. Since the baseline need not be exact, this also opens presents an opportunity for work with higher-dimensional systems for which pure optimization is very difficult. Visualization of geometry would not be necessary to determine the exact form of optimal gaits. 

The task of selecting a proper action range is still under investigation. In this work we had the ability to compare different values of this parameter and found the best one for the given robot and environment, and the interpretation of this parameter will certainly vary for other systems. Real systems would not have the luxury of trying different values until finding the one that works best. Thus, a direct line of future work would be to determine whether the action range can also be guided by system geometry.

\addtolength{\textheight}{0cm}   % This command serves to balance the column lengths

%%%%%%%%%%%%%%%%%%%%%%%%%%%%%%%%%%%%%%%%%%%%%%%%%%%%%%%%%%%%%%%%%%%%%%%%%%%%%%%%

\bibliographystyle{IEEEtran}
\bibliography{IEEEexample}

\end{document}